# Medical Image Compression using Wavelet Decomposition for Prediction Method

S.M.Ramesh
Senior Lecturer, Dept. of ECE
Bannari Amman Institute of Technology
Erode, India
E-mail: smrameshme@yahoo.co.in

Dr.A.Shanmugam
Professor, Dept. of ECE
Bannari Amman Institute of Technology
Erode, India
E-mail: dras_bit@yahoo.com

*Abstract*— **In this paper offers a simple and lossless compression method for compression of medical images. Method is based on wavelet decomposition of the medical images followed by the correlation analysis of coefficients. The correlation analyses are the basis of prediction equation for each sub band. Predictor variable selection is performed through coefficient graphic method to avoid multicollinearity problem and to achieve high prediction accuracy and compression rate. The method is applied on MRI and CT images. Results show that the proposed approach gives a high compression rate for MRI and CT images comparing with state of the art methods.**

I. INTRODUCTION

Image compression is required to minimize the storage space and reduction of transmission cost. Medical images like MRI and CT are Special images require lossless compression as a minor loss can cause adverse effects. Prediction is one of the techniques to achieve high compression. It means to estimate current data from already known data [1].

The advance image compression techniques for medical images are JPEG 2000[2] which combines integer wavelet transform with Embedded Block Coding with Optimized Truncation (EBCOT). It is an compression rate. Context based adaptive compression rate. Context based adaptive advanced technique which provides high lossless image codec (CALIC) is offered by Wu and Memon [3]. They utilized the prediction in the original CALIC but offered inter band prediction technique for remotely sensed images. A better technique for lower quality ultrasound images is offered by Przelaskrwski [4] to achieve a high compression rate. Buccigrossi and Simoncelli [5] made a statistical model and used conditional probabilities for prediction. That is a lossy method called Embedded Predictive Wavelet Image Coder(EPWIC). Yao-Tien Chen & Din-Chang Tseng proposed the Wavelet-based Medical Image Compression with Adaptive Prediction (WCAP).They used correlation analysis of wavelet coefficients to identify the basis function and further for prediction. They used lifting integer wavelet scheme for image decomposition. It is a lossless scheme to achieve highest bit rate per pixel (bpp). In (WCAP) they used backward elimination method for predictor variable selection and quantized the prediction error into Three levels (-1, 0, 1) to achieve the higher compression rate. They used DPCM for coarse bands and finally used adaptive arithmetic coding.

To achieve a high compression rate for medical images we propose wavelet based compression scheme using prediction, "Medical Image Compression using wavelet decomposition for Prediction method". The scheme uses the correlation analysis of wavelet coefficients like WCAP but adds simplicity and accuracy by excluding the requirement of selection of basis function and quantization of prediction error in coarse bands. A simple, graphic method for variable selection is introduced. The proposed scheme block diagram shown in figure.1, consists of six major stages including, image decomposition, correlation analysis of wavelet coefficients, development of prediction equation for each sub band, predictor variable selection using graphic method, arithmetic coding and reconstruction of original image. The remaining sections of this paper is organized as following, section-2 covers the Lifting Wavelet Transform of group of similar images, predictor variable selection in section- 3, experiments, results and discussion in section 4 and conclusion in the final.

II. LIFTING WAVELET TRANSFORM OF A GROUP OF IMAGES AND CORRELATION ANALYSIS

The wavelet transform is a very useful technique for image analysis and Lifting Wavelet Transform is an advance form of wavelet transform which allows easy computation, better reconstruction of original image and close approximation of some data sets. The inter scale and intra scale dependencies of wavelet coefficients are exploited to find the predictor variable. All coefficients of current, parent and aunt sub bands of each processing coefficient are found.

The correlation coefficient is based on variance and co-variance. Covariance is always measured between two matrices or dimensions while the variance is measured for a dimension with itself. The formulae for variance and covariance are as following.

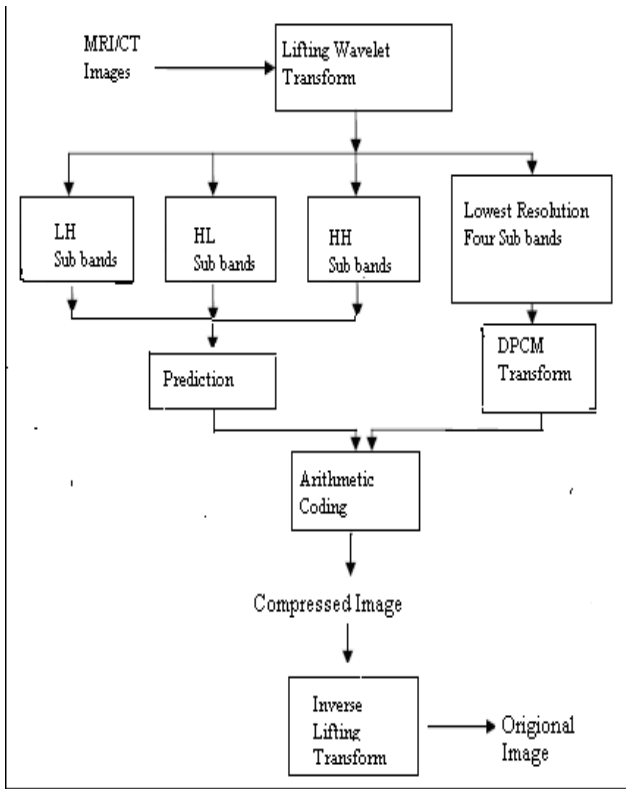

Figure1. Block diagram of proposed scheme

$$\text{Var}(X) = \frac{\sum_{i=1}^{n}(X_i - X^b)}{n-1} \quad (1)$$

$$\text{Var}(Y) = \frac{\sum_{i=1}^{n}(Y_i - Y^b)}{n-1} \quad (2)$$

$$\text{Cov}(X, Y) = \frac{\sum_{i=1}^{n}(X_i - X^b)(Y_i - Y^b)}{n-1} \quad (3)$$

$$R_{XY} = \frac{\text{Cov}(X, Y)}{\text{Var}(X)\,\text{Var}(X)} \quad (4)$$

Where $X^b$ & $Y^b$ are means of X and Y and R is correlation coefficient.

### III. PREDICTION AND PREDICTOR VARIABLE SELECTION

Our predictor is based on the linear prediction model containing k independent variables, can be written as [1]…

$$y = a_1x_1 + a_2x_2 + \cdots + a_kx_k \quad (5)$$

In this equation y is dependent variable and $x_1$, $x_2$…..$x_k$ are independent predictor variables. Where as $a_1$, $a_2$, …$a_k$ are predictor model parameters. To avoid the multicollinearity problem the number of predictor variables should be reduced. There are multiple methods to reduce the predictor variables. The best method is one which gives accurate prediction.

In the proposed method we use coefficient graphic method for selection of prediction variables. It is a simple method in which predicted and original coefficients of a sub bands are plotted for comparison. Different combination of variables is tested to select the combination which best matches the original sub band coefficient graph. This is a simple and easy method.

The sequence of prediction is from course sub band to fine sub band and from left up coefficient to the right down coefficient. The fine sub band coefficients are predicted from coarse sub band coefficients and coarse sub band coefficients are not predicted. The course sub band coefficients are than processed by Differential Pulse Code Modulation (DPCM), which is most common predictive quantization method. This method exploits correlation between successive samples of source signals and encoding based on the redundancy in sample values to give lower bit rate. This method encodes the prediction error between the sample value and its predicted value to give high compression ratio. The coarse and fine sub band coefficients are than arithmetically encoded.

### IV. EXPERIMENTS, RESULTS AND DISCUSSION

Two MRI and two CT gray scale standard test images as shown in figure 2 of size128*128 have been taken from world wide web for experiments and comparisons. MATLAB 7.0 has been used for the implementation of the proposed approach and results have been conducted on Pentium-1V, 3.20 GHz processor with a memory of 512 MB. BPP (Bits Per Pixel) metric is evaluated to compile compression result. Every image was decomposed into three scales with 10 wavelet sub bands.

Eleven correlation coefficients to the dependent c are selected which are Parent, Parent-East, Parent-West, Parent-South, Parent-North, North, North-East, North-West, West, Aunt 1 and Aunt 2. The prediction equations for coefficients for different sub bands are derived one by one. Using the coefficient graphic method, prediction variables are selected for each sub band to get accurate prediction. The compression rates for the 4 medical images using the proposed, "MICWDP" method with two famous lossless methods: SPHIT and JPEG2000 is shown in Table1. Due to proper selection of predictor Variables, proposed approach almost achieves the highest compression rates. The comparison of average encoding / decoding time of two lossless compression methods is also shown in Table 2. The proposed method makes use of "coefficient graphic method", approach successfully on medical images to get the best results.

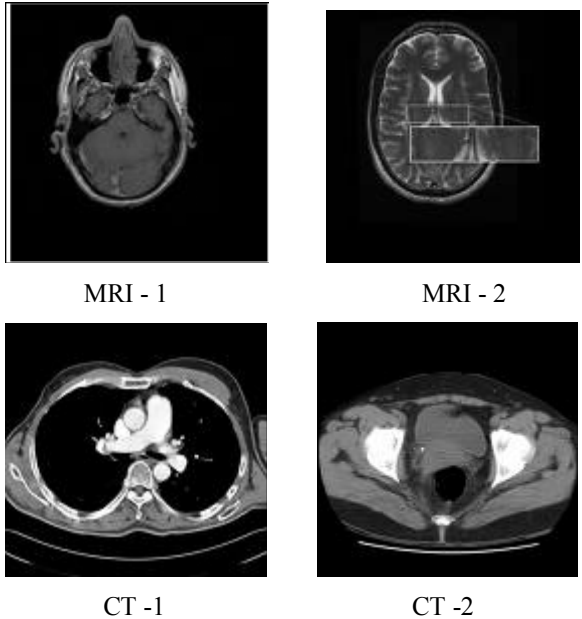

Figure. 2 2*MRI and 2*CT images taken for experiment

TABLE I
COMPARISON OF COMPRESSION RATE IN BITS/PIXEL OF DIFFERENT METHODS WITH PROPOSED METHOD

| Type | Method | | |
|---|---|---|---|
| | SPHIT | JPEG 2000 | Proposed |
| MRI -1 | 2.53 | 2.42 | 1.45 |
| MRI -2 | 3.11 | 3.12 | 1.51 |
| MRI Average | 2.82 | 2.77 | 1.48 |
| CT -1 | 1.45 | 1.32 | 1.41 |
| CT -2 | 1.79 | 1.82 | 1.43 |
| CT Average | 1.62 | 1.57 | 1.42 |

TABLE II
COMPARISON OF ENCODING/DECODING TIME OF DIFFERENT COMPRESSION METHOD

| Type | Method | | |
|---|---|---|---|
| | SPHIT | JPEG 2000 | Proposed |
| MRI Average | 1.7 / 1.9 | 0.8 / 0.8 | 2.54 / 3.13 |
| CT Average | 2.4 / 2.8 | 1.2 / 1.0 | 2.60 / 3.15 |
| Average | 2.05 / 2.35 | 2.00 / 0.9 | 2.57 / 3.14 |

## V. CONCLUSIONS

In the proposed MICWDP approach, compression rate has been improved by exploiting dependencies among wavelet coefficients [1]. A new method, i.e coefficient Graphic Method is used to avoid ulticollinearity problem which is the main contribution of this method. Comparing with the SPHIT, JPEG2000 and proposed achieves the highest compression rate.

## REFERENCES

[1] Yao-Tien Chen and Din-Chang Tseng, Wavelet-based medical Image compression with adaptive prediction. In: proc,International symposium on Intelligent Signal Processing and Communication Systems, December 2005-Hong Kong p.825-8 and Computerized medical Imaging and graphics 31(2007) 1-8

[2] Krishnan K.marcellin MW, Bilgin A, Nadar M.Prioritization of compressed data by tissue type using JPEG2000, In:proc.SPIE medical imaging 2005-PACS and imaging informatics.2005. p.181-9

[3] Wu X, Memon N. Context-based, adaptive, lossless Image coding. IEEE Trans Image Process 1997;6(5):656-64.

[4] Przelaskowski A. lossless encoding of medical images:hybrid modification of statistical modeling-based conception. J Electron Imaging 2001;10(4):966-76.

[5] Bussigrossi RW, Simoncelli EP. Image compression via joint Statistical characterization in the wavelet domain. IEEE Trans Image Process 1999;8(12);1688-701.

[6] Rafael C. Gonzalez and Richard E. Woods, Digital Image Processing, 2nd Edition, Printice Hall Inc, 2002.

[7] Ian Kaplan, Basic lifting scheme wavelets, February 2002(revised)

[8] Lindsay I Smith, A tutorial on principal component analysis, 26 February 2002.

[9] Majid Rabbani and Paul W. Jones, Image compression techniques.

[10] In H. Witten, Radford M. Neal and John G. Cleary, Arithmetic Coding for data compression.

[11] Mark Nelson, Arithmetic coding, Dr. Dobbs Journal,February, 1991.

## AUTHORS PROFILE

**Dr.A.Shanmugam** received the B.E, degree in Electronics and Communication Engineering from PSG College of Technology., Coimbatore, Madras University, India in the year 1972 and the M.E, degree in Applied Electronics from College of Engineering, Guindy, Chennai, Madras University, India in the year 1978 and received the Ph.D. in Computer Networks from PSG College of Technology., Coimbatore, Bharathiyar University, India in the year 1994.From 1972 to 1976, he served as a Testing Engineer at Test and Development Center, Chennai, India. From 1978 to 1979, he served as a Lecturer in the Department of Electrical Engineering, Annamalai University, India. From 1979 to 2002, he served different level as a Lecturer, Asst.Professor, Professor and Head in the Department of Electronics and Communication Engineering of PSG College of Technology, Coimbatore, India. Since April 2004,he assumed charge as the Principal, Bannari Amman Institute of Technology, Sathyamangalam, Erode, India. He works in field of Optical Networks, broad band computer networks and wireless networks, Signal processing specializing particularly in inverse problems, sparse representations, and over-complete transforms.

Dr.A.Shanmugam received "Best Project Guide Award" five times from Tamil Nadu state Government. He is also the recipient of "Best Outstanding Fellow Corporate Member Award" by Institution of Engineers (IE),India - 2004 and "Jewel of India" Award by International Institute of Education and Management, New Delhi–2004 and "Bharatiya Vidya Bhavan National Award for Best Engineering College Principal 2005" by Indian Society for Technical Education (ISTE). "Education Excellence Award" by All India Business& Community Foundation, New Delhi.

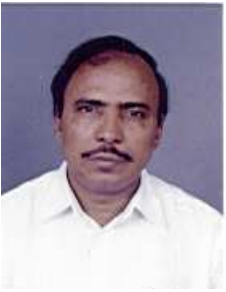 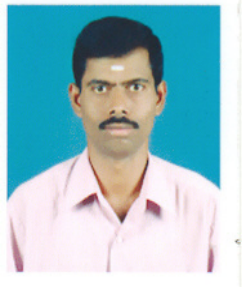

Dr.A.Shanmugam          Mr.S.M.Ramesh

**S.M.Ramesh** received the B.E degree in Electronics and Communication Engineering from National Institute of Technology (Formerly Regional Engineering College), Trichy, Bharathidhasan University, India in the year 2001 and the M.E, degree in Applied Electronics from RVS College of Engineering and Technology, Dindugal, Anna University, India in the year 2004. From 2004 to 2005, he served as a Lecturer in the Department of Electronics and Communication Engineering, Maharaja Engineering College, Coimbatore, India. From 2005 to 2006, he served as a Lecturer in the Department of Electronics and Communication Engineering, Nandha Engineering College, Erode, India. Since June 2006, he served as Sr.Lecturer, in the Department of Electronics and Communication Engineering Bannari Amman Institute of Technology, Sathyamangalam, and Erode, India. He is currently pursuing the Ph.D. degree, working closely with Prof. Dr.A.Shanmugam and Prof Dr.R.Harikumar.